\pdfoutput=1

\documentclass[11pt]{article}

\usepackage[preprint]{acl}

\usepackage{times}
\usepackage{latexsym}

\usepackage[T1]{fontenc}

\usepackage[utf8]{inputenc}

\usepackage{microtype}

\usepackage{inconsolata}

\usepackage{graphicx}
\usepackage{booktabs}
\usepackage{algorithm}
\usepackage{algpseudocode}
\usepackage{hyperref}
\usepackage{subcaption}
\usepackage{amsfonts}
\usepackage{multirow}
\usepackage{amsmath}
\usepackage{pifont}
\usepackage[normalem]{ulem}
\useunder{\uline}{\ul}{}
\usepackage{listings}
\usepackage{tcolorbox}
\usepackage{tabularx}
\usepackage{makecell}
\usepackage{xspace}
\usepackage{adjustbox}
\usepackage{dirtree}

\newcommand{\fwname}{\textsc{PEFT-Factory}\xspace}

\title{\fwname: Unified Parameter-Efficient Fine-Tuning of Autoregressive Large Language Models}

\author{Robert Belanec$^{\spadesuit}$$^\dagger$, Ivan Srba$^\dagger$, Maria Bielikova$^\dagger$ \\
$^{\spadesuit}$ Faculty of Information Technology, Brno University of Technology, Brno, Czechia \\
$^\dagger$ Kempelen Institute of Intelligent Technologies, Bratislava, Slovakia\\
\texttt{\{name.surname\}}@kinit.sk\\ 
Demonstration video: {\bf\url{https://youtu.be/Q3kxvlyO-XY}}\\
Installation package: {\bf\url{https://pypi.org/project/peftfactory}}\\
Live demo: {\bf\url{https://peftfactory.kinit.sk}}
}

\begin{document}
\maketitle
\begin{abstract}
Parameter-Efficient Fine-Tuning (PEFT) methods address the increasing size of Large Language Models (LLMs). Currently, many newly introduced PEFT methods are challenging to replicate, deploy, or compare with one another. To address this, we introduce \fwname, a unified framework for efficient fine-tuning LLMs using both off-the-shelf and custom PEFT methods. While its modular design supports extensibility, it natively provides a representative set of 19 PEFT methods, 27 classification and text generation datasets addressing 12 tasks, and both standard and PEFT-specific evaluation metrics. As a result, \fwname provides a ready-to-use, controlled, and stable environment, improving replicability and benchmarking of PEFT methods. \fwname is a downstream framework that originates from the popular LLaMA-Factory, and is publicly available at \url{https://github.com/kinit-sk/PEFT-Factory}.
\end{abstract}

\section{Introduction}
Large Language Models (LLMs) \citep{minaee2024large, radford2019language, dubey2024llama, raffel2020exploring} achieved remarkable results in many different Natural Language Processing (NLP) tasks mainly after the introduction of the transformer architecture \citep{vaswani2017attention}. However, for its great scaling capabilities \citep{kaplan2020scaling}, the size of the model in terms of trainable parameters is continuously increasing accordingly. The growing number of LLM parameters rendered fine-tuning computationally expensive, data-hungry, and hardly accessible for many researchers and practitioners. 

Parameter-Efficient Fine-Tuning (PEFT) methods \citep{xu2023parameter, ding2023parameter, lialin2023scaling, han2024parameterefficient} aim to address these issues by training only a small percentage of the full model's parameters while achieving performance comparable to that of full model fine-tuning. Such a decrease in trainable parameters can be achieved by adding new parameters \citep{houlsby2019parameter}, selecting specific parameters \citep{ben-zaken-etal-2022-bitfit} for training, or by reparameterizing the model with a smaller number of parameters \citep{hu2022lora}.

Due to their effectiveness, PEFT methods have gained popularity and become an attractive research area, with many new contributions being introduced each year \citep{han2024parameterefficient}. However, the large number of newly introduced PEFT methods makes it harder to compare their contributions, resulting in only a few established PEFT methods (LoRA variants in most cases) being used in practice, while others, which may be more effective, remain largely unused. Moreover, many new PEFT methods lack a fully functional open-source implementation, and essential details on the experimental setup often prevent fellow researchers from replicating their results. Therefore, many researchers \citep{asai-etal-2022-attempt, shi2024dept, tang2025adept} have to rely on reported performance metrics, while there is a risk of not replicating exactly the same experimental setup for their own methods (making a comparison potentially unfair), as well as it is not feasible to rerun the existing solutions on additional datasets/tasks. Lastly, many established PEFT methods lack proper evaluation on autoregressive LLMs.

To tackle these accumulating problems, \textbf{we introduce \fwname, an easy-to-use and modular framework for efficient fine-tuning and evaluation of LLMs using different PEFT methods}. 
\fwname is based on the popular and open-source fine-tuning framework \textit{LLaMA-Factory}~\citep{zheng2024llamafactory}. It is built using PyTorch~\citep{pytorch} and utilizes open-source Python modules for training LLMs, including Transformers~\citep{wolf-etal-2020-transformers}, PEFT~\citep{peft}, TRL~\citep{vonwerra2022trl}, and Adapters~\citep{poth-etal-2023-adapters}.

Our main contributions are as follows:
\begin{itemize}
    \item \fwname provides a support for off-the-shelf methods from popular PEFT provider frameworks like \textit{HuggingFace PEFT} \citep{peft} or \textit{Adapters} \citep{poth-etal-2023-adapters} as well as dynamic loading of \textit{custom user-created PEFT methods}. In contrast to the existing solutions, it provides so-far-missing support for \textit{soft prompt-based}, \textit{adapter-based} and \textit{selective} PEFT methods; as well as for \textit{classification} tasks.
    \item \fwname natively provides a representative set of \textit{19 PEFT methods}, \textit{27 classification and text generation datasets} addressing \textit{12 unique tasks}, and standard as well as PEFT-specific \textit{evaluation metrics}. This ready-to-use setup enables quick adoption and experimentation by researchers and practitioners, significantly improving the currently limited replicability and benchmarking of PEFT methods.
    \item \fwname is designed with future \textit{extensibility} in mind and provides a fully open-source codebase for anyone to use. It implements a standardized PEFT interface to enable modular addition of newly created PEFT methods. Similarly, it allows easy extension for additional datasets.
\end{itemize}

\begin{table*}[t!]
\centering
\resizebox{\textwidth}{!}{%
\begin{tabular}{@{}l|cccc|ccc@{}}
\toprule
 & Reparametrized & Soft Prompt-Based & Adapter-Based & Selective & Classification Datasets & Classification Metrics & Extensibility \\ \midrule
\textbf{\fwname} & 8 & 5 & 4 & 2 & \ding{52}  & \ding{52}  & datasets, models, PEFT methods  \\
LLaMA-Factory & 7 & 0 & 0 & 0 & \ding{56} & \ding{56} & datasets, models \\
FastChat & 3 & 0 & 0 & 0 & \ding{56} & \ding{56} & datasets, models \\
LitGPT & 2 & 0 & 0 & 0 & \ding{52}  & \ding{52}  & datasets, models \\
LMFlow & 3 & 0 & 0 & 0 & \ding{56}  & \ding{56}  & datasets, models  \\
Axolotl & 2 & 0 & 0 & 0 & \ding{56} & \ding{56} & datasets, models \\
Open-Instruct & 3 & 0 & 0 & 0 & \ding{56} & \ding{56} & \ding{56} \\
H2O LLM Studio & 3 & 0 & 0 & 0 &  \ding{52}  &  \ding{52}  & datasets, models  \\
GPT4All & 2 & 0 & 0 & 0 & \ding{56} & \ding{56} & \ding{56} \\ \bottomrule
\end{tabular}%
}
\caption{Comparison of \fwname to popular fine-tuning frameworks. Only \fwname allows for out-of-the-box non-reparametrization efficient fine-tuning with the extensibility of additional and custom fine-tuning methods. Comparison at the level of individual PEFT methods can be found in Table \ref{app:tab:peft_comparison} of Appendix \ref{app:sec:details}.}
\label{tab:comparison}
\end{table*}

\section{Related Work}
There has been a significant rise in the number of frameworks used for training of (not only) LLMs. 
LLaMA-Factory~\citep{zheng2024llamafactory} is a recent addition to such frameworks and offers end-to-end and easy-to-use training of LLMs ranging across all of the stages (from pre-training to alignment via reinforcement learning). LLaMA-Factory also provides a graphical user interface called LLaMABoard, implemented in Gradio~\citep{abid2019gradio}, which enhances the ease of use of LLaMA-Factory. Despite being a really popular and useful tool for LLM training, LLaMA-Factory still provides fine-tuning only with LoRA~\citep{hu2022lora} and its variants, namely QLoRA~\citep{qlora}, DoRA~\citep{dora}, LoRA+~\citep{lora+}, PiSSA~\citep{pissa}, and GaLore~\citep{galore}. With the recent update, LLaMA-Factory also allows Orthogonal Fine-Tuning (OFT)~\citep{qiu2023controlling}, which utilizes the Cayley transformation~\citep{cayley1846quelques} to fine-tune only orthogonal vectors. Nevertheless, the selection of PEFT methods in LLaMA-Factory still remains limited. Lastly, LLaMA-Factory primarily focuses on text-generation problems and does not incorporate the possibility of casting text-generation problems as classification tasks. Our framework \fwname addresses both the limited number of available PEFT methods and the potential for fine-tuning LLMs for classification.

There are also other LLM training frameworks that are less easy to run (compared to LLaMA-Factory) and have their specific benefits. FastChat~\citep{zheng2023judging} is a specialized framework for training LLMs for chat-completion. LitGPT~\citep{litgpt-2023} and LMFlow~\citep{diao2023lmflow} are extensible and convenient general training frameworks that support various generative models and training methods. Axolotl~\citep{axolotl} is a terminal-based tool for efficient post-training of LLMs without sacrificing functionality or scale. Open-Instruct~\citep{wang2023far} focuses on instruction fine-tuning for LLMs and provides multiple models and recipes for this purpose. H2O LLM Studio\footnote{\url{https://github.com/h2oai}} is a more enterprise-oriented, all-in-one tool that also provides a graphical interface for developing and deploying LLM models. GPT4All~\citep{gpt4all} creates a user-friendly interface around llamacpp. ColossalAI~\citep{colossalai} focuses on delivering a framework for distributed fine-tuning.

In addition, LLaMA-Adapter~\citep{zhang2023llamaadapter} and LLaMA-Accesory~\citep{gao2023llamaadapterv2} are more lightweight frameworks, where LLaMA-Adapter adds trainable adapters to (not only) LLaMA models and LLaMA-Accesory provides a full toolkit for LLM development. LLaMA-Adapter is often implemented in previously-mentioned frameworks, such as LitGPT.Table \ref{tab:comparison} provides a summary of unique \fwname features when compared with popular fine-tuning frameworks as well as our upstream framework LLaMA-Factory. Based on our analysis of related frameworks and to the best of our knowledge, we have identified 3 key features that are currently missing or limited, and are novel in our work: 1) training of LLMs with other than reprametrization-based PEFT methods, 2) modular and dynamic addition of new PEFT methods, and 3) support for training and evaluation of LLMs for classification.

\section{\fwname}
\begin{figure*}
  \centering
  \includegraphics[width=0.9\textwidth]{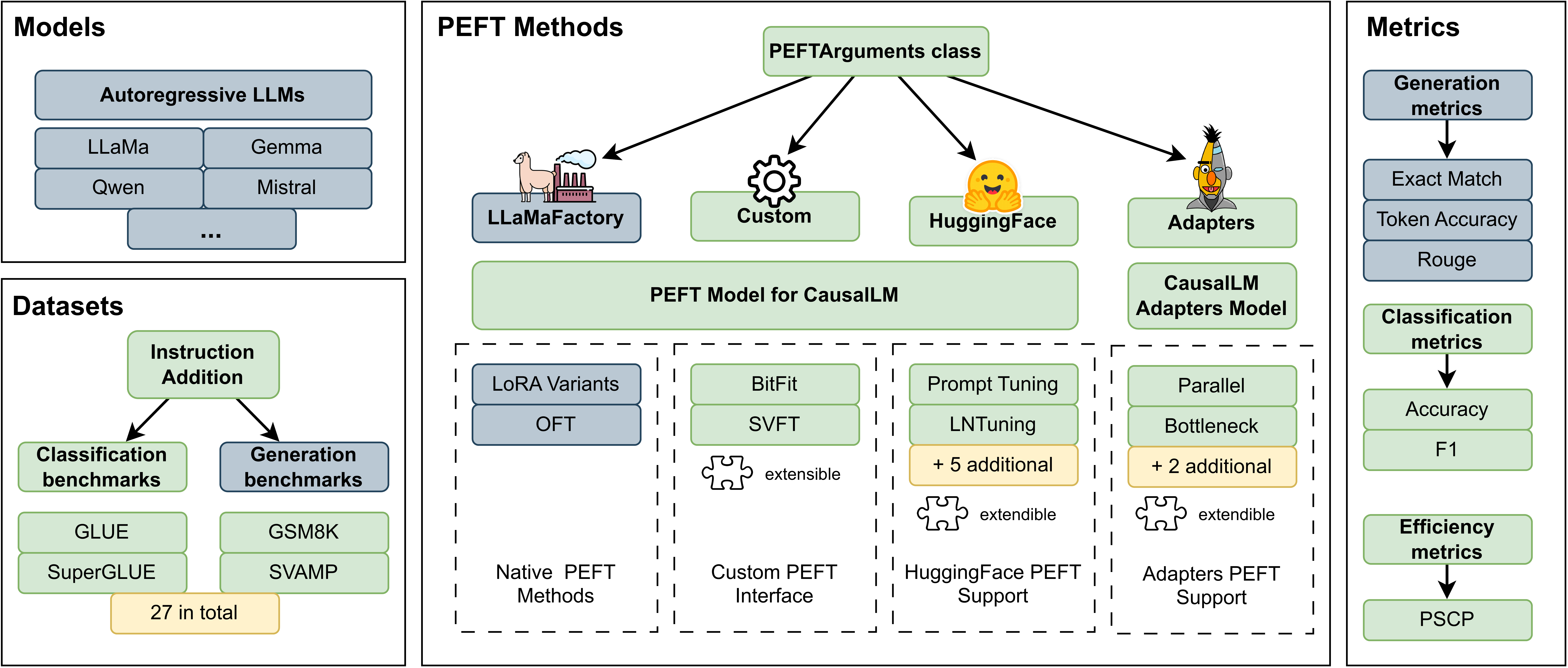}
  \caption{Diagram representing the components of \fwname. The four main overarching components of \fwname are PEFT Methods, Datasets, Models, and Metrics, which are further defined by their subcomponents. Components represented by green color are implemented in \fwname, components in blue color are native to LLaMA-Factory~\citep{zheng-etal-2024-llamafactory}. Additionally, the Adapters library requires a different model class than the rest of the PEFT provider frameworks.}
  \label{fig:peftfactory}
\end{figure*}

The \fwname consists of four main components: 1) PEFT Methods, 2) Datasets, 3) Models, and 4) Metrics, as also depicted in Figure \ref{fig:peftfactory}.

In the \textit{PEFT methods} component, we design and implement support for reparameterized, soft prompt-based, adapter-based, and selective PEFT methods, from HuggingFace PEFT~\citep{peft} and Adapters~\citep{poth-etal-2023-adapters} PEFT provider frameworks. We also provide a custom PEFT interface for more advanced users to provide and dynamically load their custom PEFT methods into \fwname. Currently, we include \textit{19 different PEFT methods} (out of them, 7 are natively provided by the LLaMA-Factory). Full listing of PEFT methods covered by \fwname can be found in Table \ref{app:tab:peft_comparison} of Appendix \ref{app:sec:details}.

The core of the \textit{Datasets} component is the dataset loader supporting datasets from classification tasks, with the possibility of adding separate instructions for instruction fine-tuned models (a missing feature of LLaMA-Factory). Additionally, we include and adapt multiple well-known classification benchmarks, as well as text-generation tasks, totalling \textit{27 datasets}.

Regarding \textit{Models}, \fwname leverages the existing support provided by LLaMA-Factory. It enables the utilization of a wide range of models from different model families, spanning from 0.5 (e.g., Qwen 2.5~\citep{Yang2024Qwen25TR}) to 671 (e.g., DeepSeek R1~\citep{guo2025deepseek}) billion parameters. For demonstration purposes, we selected Llama-3.2-1B-Instruct~\cite{dubey2024llama} as it is a popular representative of a reasonable size that allows fast training to demonstrate \fwname. 

Within the \textit{Metrics} component, we add classification and performance-based metrics into the evaluation of LLMs trained using PEFT methods. This includes the addition of standard classification metrics, such as accuracy and F1, as well as the PSCP metric~\citep{peftbench}, which incorporates various efficiency factors into the results.

\subsection{Off-The-Shelf PEFT Support}\label{sec:off-the-shelf}
There are many different PEFT methods included off-the-shelf within the PEFT provider libraries, such as HuggingFace PEFT and Adapters. In the current state, we include 10 different off-the-shelf PEFT methods that we have tested with different state-of-the-art LLMs, namely, from the \textbf{Adapters library} -- \textit{Parallel Adapter}~\citep{he2022towards}, \textit{Bottleneck Adapter}~\citep{houlsby2019parameter}, and \textit{Sequential Bottleneck Adapter}~\citep{pfeiffer-etal-2020-mad}; and from the \textbf{HuggingFace PEFT} -- \textit{Prompt Tuning}~\citep{lester-etal-2021-power}, \textit{Prefix Tuning}~\citep{li-liang-2021-prefix}, \textit{P-Tuning}~\citep{liu2023gpt}, \textit{P-Tuningv2}~\citep{liu-etal-2022-p}, \textit{MTP}~\citep{wang2023multitask}, \textit{LNTuning}~\citep{zhao2024tuning}, and \textit{IA$^3$}~\citep{liu2022few}. Moreover, \fwname enables to easily \textit{add more of such off-the-shelf PEFT methods simply by updating two constants in the code}.

From the implementation perspective, to include support for these libraries in \fwname, we created a unified \texttt{PeftArguments} class that inherits both the \texttt{PEFTConfig} and \texttt{AdapterConfig} classes for typing purposes. This joint class is then used for parsing the parameters from configurations via \texttt{HFArgumentParser}. We store all supported off-the-shelf PEFT methods in several constants, specifically lists \texttt{HF\_PEFT\_METHODS} and \texttt{ADAPTERS\_METHODS}, along with their counterpart mapping dictionary constants \texttt{PEFT\_CONFIG\_MAPPING} and \texttt{ADAPTERS\_CONFIG\_MAPPING}. If the PEFT method in the configuration is contained within the mapping dictionaries, a specific config is used. Otherwise, it will default to \texttt{PeftArguments} (in file \texttt{hparams/parser.py}, function \texttt{\_parse\_train\_args}). Importantly, every PEFT method, whether added from the Adapters library or the HuggingFace PEFT library, comes with its own set of parameters or hyperparameters that can be tuned. \fwname automatically detects the parameters required by a specific PEFT method and uses the \texttt{HFArgumentParser} to parse them from config files (if parameters are not specified in the config, default values are used from the original implementation). This allows easy configuration and hyperparameter tuning via a YAML config file or the Gradio user interface.

The newly created and parsed \texttt{PeftArguments} are then forwarded to the \texttt{model loader}, where the \texttt{init\_adapter} method creates the PEFT model (in file \texttt{model/model\_utils/adapter.py}, functions \texttt{\_setup\_custom\_peft}, \texttt{\_setup\_adapters\_peft} and \texttt{\_setup\_hf\_peft}). From this part, we leave the loading, training, and model saving on PEFT libraries and the LLaMA-Factory framework.

\subsection{Custom PEFT Interface}\label{sec:custom}

In contrast to LLaMA-Factory, \fwname implements a dynamic loading mechanism for custom PEFT methods, ultimately enabling its extensibility and modularity. This design allows researchers and practitioners to seamlessly integrate custom PEFT implementations without modifying the core codebase of \fwname. To demonstrate our Custom PEFT Interface, we replicated 2 PEFT methods that are not a part of any off-the-shelf PEFT framework, namely BitFit~\citep{ben-zaken-etal-2022-bitfit} and SVFT~\citep{lingam2024svft} (located in the \href{https://github.com/kinit-sk/PEFT-Factory/tree/main/peft}{\texttt{peft directory}}).

During the process of dynamic loading, the \texttt{peft\_loader} module automatically discovers and loads PEFT methods from a structured directory hierarchy (in file \texttt{extras/peft\_loader.py}, function \texttt{discover\_custom\_peft\_methods}). Each custom PEFT method is organized in its own subdirectory containing two essential components: a \texttt{config.py} file defining a \texttt{PeftConfig} subclass, and a \texttt{model.py} file implementing a \texttt{BaseTuner} subclass. The configuration and model subclasses need to inherit the \texttt{PeftConfig} dataclass and \texttt{BaseTuner} abstract class from the HuggingFace PEFT library. The methods required to be implemented are then specified by the \texttt{BaseTuner} description in \texttt{\href{https://github.com/huggingface/peft/blob/main/src/peft/tuners/tuners_utils.py}{tuner\_utils.py}}.

The loader validates each implementation by checking for required attributes (\texttt{peft\_type} for configurations and \texttt{prefix} for model classes) before registration. The loader dynamically loads the config and model subclasses, registering them via the \texttt{register\_peft\_method} function (in file \texttt{peft\_loader.py}), which adds the config and model to the constants of the Hugging Face PEFT library. Additionally, this process is also defined by the Algorithm \ref{app:alg:disc}, which explains how dynamic loading is implemented.

\begin{algorithm}[ht]
\caption{Dynamic PEFT Method Discovery}
\label{app:alg:disc}
\begin{algorithmic}[1]
\State \textbf{Input:} PEFT directory path $D$
\State \textbf{Output:} Dictionary $M$ mapping method names to (config, model) tuples
\State $M \gets \emptyset$
\For{each subdirectory $d$ in $D$}
    \If{$d$ contains \texttt{config.py} and \texttt{model.py}}
        \State Load config class $C$ from \texttt{config.py}
        \State Load model class $T$ from \texttt{model.py}
        \If{$C$ validates and $T$ validates}
            \State $M[\text{name}(d)] \gets (C, T)$
        \EndIf
    \EndIf
\EndFor
\end{algorithmic}
\end{algorithm}

After the dynamic loading, the corresponding custom PEFT method is handled similarly to off-the-shelf methods as described in Section \ref{sec:off-the-shelf}. As a result, the ease of configuration and hyperparameter tuning for the newly added PEFT methods also remains unchanged.

To add a new method, it is required to match the directory structure inside the \textit{PEFT methods directory} (the directory can be specified by the environment variable \texttt{PEFT\_DIR} with \texttt{./peft} as the default directory) to match the organizational structure and class inheritance. We provide information about example method templates in Appendix \ref{app:sec:template} as well as in the \fwname documentation\footnote{For detailed information, please visit the \href{https://peft-factory.readthedocs.io/en/latest/contributing.html\#adding-peft-methods}{Adding PEFT Methods section} of the \fwname documentation.}.

This plugin-style architecture promotes code reusability and enables fast prototyping of novel PEFT methods. Researchers can develop and test new methods separately, with the framework automatically integrating them at runtime.

The dynamic loading approach has proven particularly valuable for comparative studies, allowing researchers to evaluate multiple PEFT variants under identical experimental conditions without code duplication or version control conflicts.

\subsection{Improved Dataset Loader}
Besides a native support of text generation tasks (inherited from the LLaMA-Factory), we add support for classification tasks (in case of autoregressive models, the classification task $Pr_\theta(y|X)$ is cast as a generation $Pr_\theta(Y|X)$ task). To this end, we adapt and include multiple classification benchmarks, including GLUE~\citep{wang2018glue} and SuperGLUE~\citep{wang2019superglue}. This was largely possible due to the dataset loading feature included in LLaMA-Factory. However, we needed to enhance the dataset loader with an additional parameter called \textit{instruction}. This optional attribute was added to the \textit{DatasetAttr} class (in file \texttt{data/processor/parser.py}), and during data preprocessing, the instruction is prepended to the input text for the LLM (in file \texttt{data/processor/converter.py}, class AlpacaDatasetConverter). This allows adding instructions that have dataset-specific formatting (e.g., ones recommended by the dataset authors)\footnote{For detailed information, please visit the \href{https://peft-factory.readthedocs.io/en/latest/contributing.html\#adding-datasets}{Adding Datasets section} of the \fwname documentation.} or are designed for instruction tuning tasks.

\paragraph{Addition of classification datasets.} From the GLUE benchmark, we include 8 classification datasets separated into 6 tasks, namely \textbf{natural language inference (NLI)} -- \textit{MNLI}~\citep{williams-etal-2018-broad}, \textit{QNLI}~\citep{rajpurkar2016squad}, \textit{RTE}~\citep{dagan2006pascal,bar2006second,giampiccolo2007third,bentivogli2009fifth}; \textbf{paraphrase classification} -- \textit{QQP}~\footnote{\href{https://quoradata.quora.com/First-Quora-Dataset-Release-Question-Pairs}{https://quoradata.quora.com/First-Quora-Dataset-Release-Question-Pairs}}, \textit{MRPC}~\citep{dolan2005automatically}; \textbf{sentiment classification} -- \textit{SST-2}~\citep{socher2013recursive}; \textbf{sentence similarity} -- \textit{STS-B}~\citep{cer-etal-2017-semeval} and \textbf{acceptability classification} -- \textit{CoLA}~\citep{warstadt-etal-2019-neural}.

From SuperGLUE, we include 7 datasets separated into 4 tasks, namely \textbf{natural language inference (NLI)} -- \textit{CB}~\citep{de2019commitmentbank}; \textbf{question answering} -- \textit{MultiRC}~\citep{khashabi2018looking}, \textit{ReCoRD}~\citep{zhang2018record}, \textit{BoolQ}~\citep{clark-etal-2019-boolq}, \textit{COPA}~\citep{roemmele2011choice}; \textbf{word sense disambiguation} -- \textit{WiC}~\citep{pilehvar-camacho-collados-2019-wic} and \textbf{coreference resolution} -- \textit{WSC}~\citep{levesque2011winograd}.

\paragraph{Addition of generation datasets.} We also include generation datasets that are commonly used to benchmark generative LLMs. We cover 6 datasets for reasoning and natural language understanding separated into 3 tasks, namely \textbf{question answering} -- \textit{MMLU}~\citep{hendrycks2021measuring}, \textit{PIQA}~\citep{bisk2020piqa}, \textit{SIQA}~\citep{sap-etal-2019-social}, \textit{OBQA}~\citep{khot-etal-2019-whats}; \textbf{natural language inference (NLI)} -- \textit{HellaSwag}~\citep{zellers-etal-2019-hellaswag}; \textbf{commonsense reasoning} -- \textit{WinoGrande}~\citep{sakaguchi2021winogrande}; 3 datasets for mathematical problem solving separated into 3 tasks, namely \textbf{question answering} -- \textit{MathQA}~\citep{amini-etal-2019-mathqa}; \textbf{math word problems} -- \textit{GSM8K}~\citep{cobbe2021training} and \textbf{simple math problems} -- \textit{SVAMP}~\citep{patel-etal-2021-nlp}; and 3 datasets for code generation, namely \textit{Conala}~\citep{yin2018learning}, \textit{CodeAlpacaPy}~\citep{codealpaca}, and \textit{APPS}~\citep{hendrycks2021measuring}. 

\paragraph{Adapting and preprocessing datasets.} Some datasets may require further (mostly minor) format changes to be compatible with the input formatting of \fwname. We further describe this preprocessing in the Appendix \ref{app:details:preprocess}.

\subsection{Classification and Efficiency Metrics}
\fwname also calculates classification and efficiency metrics during the prediction phase in addition to already existing token accuracy and semantic similarity metrics (i.e., Rouge~\citep{lin-2004-rouge} and Bleu~\citep{papineni-etal-2002-bleu}). From the classification metrics, \fwname implements standard Accuracy and F1 metrics. To measure efficiency during the evaluation of PEFT methods, \fwname implements the PSCP metric~\citep{peftbench}, a highly adjustable metric that considers various efficiency parameters (e.g., number of parameters, memory usage, inference time).

We implement these metrics within the \texttt{train/sft/metric.py} file for supervised fine-tuning, following the pattern from LLaMA-Factory and utilizing separate data classes for each metric. Specifically, we name the classes \textit{ComputeAccuracy}, \textit{ComputeF1}, and \textit{ComputePSCP}. For classification, we include a binary flag attribute within the training arguments, called \textit{compute\_classification\_metrics}, which enables or disables the computation of classification metrics. For the efficiency metrics, we include a binary flag \textit{compute\_pscp}. Additional information on the usage of efficiency metrics can be found in Appendix \ref{app:details:pscp}.

\section{\fwname-enabled Use Cases}
The extensibility of PEFT methods and datasets, together with a ready-to-use, controlled, and stable environment, is a key factor of \fwname that aims to promote further research on PEFT methods. To demonstrate how \fwname improves reproducibility and benchmarking of PEFT methods, we present two specific use cases.

\subsection{PEFT Methods Reproducibility}
How the modular design of \fwname promotes reproducibility and transparency of newly created PEFT methods can be seen in the use case, when fellow researchers and practitioners develop new PEFT methods. 

Currently, when a new PEFT method is developed, the published source code is often not fully functional or difficult to reproduce. In addition, the authors often have to implement code for training and evaluation of the PEFT method from scratch, which is often repetitive, increases the probability of mistakes in the code, and is prone to inconsistencies in the final results.

In our scenario, authors only need to create a minimum number of files that are directly and solely connected to the design of the PEFT method itself. If the authors maintain the structure compatible with the \fwname custom PEFT interface, they can simply share it within the PEFT methods directory, create a configuration for training and evaluation, and run experiments on vast amounts of datasets and autoregressive models. Additionally, if the authors choose to implement their method inside any of the supported PEFT provider frameworks (i.e., Hugging Face PEFT or Adapters), only a small change is needed to contribute it to the next version of the \fwname\footnote{We provide information on how to request the addition of a new PEFT provider method in the \href{https://peft-factory.readthedocs.io/en/latest/contributing.html}{Contributing page} of \fwname documentation}.

\subsection{PEFT Methods Benchmarks}
Another possible use case of \fwname is to benchmark PEFT methods. To this end, PEFT-Factory provides a standardized and reproducible environment that eliminates inconsistencies in experimental setups (e.g., different seeds, hyperparameters or dataset splits), allowing researchers to reliably compare PEFT methods under identical conditions. To illustrate the benchmarking capability, Table \ref{tab:bench} provides results from fine-tuning three PEFT methods on four different datasets using the LLaMA-3.2-1B-Instruct~\citep{dubey2024llama} autoregressive model. Even such a small demonstrative comparison would require significant code-base preparation to execute the experiments, which \fwname eliminates to a minimum. From this evaluation, we can see that BitFit achieves the highest results in most of the datasets.

As a more complex benchmarking use case, we refer to our parallel work \cite{peftbench}, in which we introduce the PEFT-Bench -- a benchmark of the efficiency of PEFT methods fully conducted within \fwname. PEFT-Bench provides the first unified, end-to-end benchmarking suite for evaluating PEFT methods on modern autoregressive LLMs, covering 27 datasets, 12 task types, and 7 diverse PEFT techniques. This benchmark was only possible due to \fwname, which serves as the underlying engine. 

\fwname thus allows the community to easily extend the PEFT-Bench with new PEFT methods or even design new benchmarks with minimal effort. By ensuring experiment equivalency, replicability, and ease of extensibility, \fwname empowers researchers and practitioners to rigorously evaluate existing PEFT approaches and accelerate the development of new ones.

\begin{table}[t]
\centering
\resizebox{0.8\columnwidth}{!}{%
\begin{tabular}{@{}l|llll@{}}
Method & SST-2 & CoLA & WSC & SVAMP \\ \midrule
BitFit & \textbf{97.5} & 86.9 & \textbf{55.2} & \textbf{92.3} \\
IA$^3$ & 95.3 & 85.3 & 3.6 & 84.1 \\
Prefix Tuning & 96.3 & \textbf{88.8} & 0.8 & 91.4
\end{tabular}%
}
\caption{Macro F1 results to demonstrate the benchmarking use case of \fwname on different datasets for different PEFT methods.}
\label{tab:bench}
\end{table}

\section{Conclusion and Future Work}\label{sec:conclusion}
We introduce \fwname, a modular and extensible framework for fine-tuning modern autoregressive models using recent and diverse PEFT methods. \fwname not only provides a way to utilize PEFT methods but also implements support for various PEFT providers and a custom PEFT interface to promote replicability and transparency when designing new PEFT methods. When comparing \fwname to various popular fine-tuning frameworks, as well as to our upstream framework, LLaMA-Factory, its novelty lies in suppo
rting different PEFT methods, classifying tasks with custom instructions, and providing PEFT- and dataset-level extensibility.

\paragraph{Sustainability and Maintenance.} To keep up with the updates included in LLaMA-Factory (which often include support of new LLMs or improvements in the training pipeline), we will regularly release a new version of \fwname (this includes merging the upstream changes into our repository). Additionally, to include new features in \fwname itself, we will reguraly release a separate version increment. Each change will be documented in the changelog of the specific release.

As the next steps, we would like to increase support for additional PEFT off-the-shelf methods, as well as reproduce some popular PEFT methods that are not currently supported by any of the PEFT provider frameworks. We believe that \fwname is an important and enabling tool that will promote the research of PEFT methods and allow their fair and consistent evaluation.

\section*{Acknowledgments}
This work was partially funded by the European Union, under the project LorAI - Low Resource Artificial Intelligence, GA No. \href{https://doi.org/10.3030/101136646}{101136646}; and by the European Union NextGenerationEU through the Recovery and Resilience Plan for Slovakia under the project No. 09I01-03-V04-00006.

Part of the research results was obtained using the computational resources procured in the national project, National Competence Centre for High Performance Computing (project code: 311070AKF2), funded by ERDF, EU Structural Funds Informatization of Society, Operational Program Integrated Infrastructure.

\bibliography{ref}

\appendix

\section{Ethical Considerations}

The experiments in this paper were conducted using publicly available datasets, including SST-2, CoLA, WSC, and SVAMP, as cited by the original authors. As we were unable to determine the licenses for all used datasets, we have opted to use them in the limited form possible, adhering to the terms of use of the GLUE and SuperGLUE benchmarks. As the datasets are commonly used in other related works and have been published in scientific works that went through an established review process, we do not check for the presence of any offensive content, as it was already removed by the authors of these publicly available datasets. Additionally, we do not collect or utilize any personally identifiable information or offensive content, and we do not engage in crowdsourcing for data annotation in any form. To our knowledge, we are not aware of any potential ethical harms or negative societal impacts of our work, apart from those related to the field of Machine Learning (i.e., the use of computational resources that consume energy and produce heat, resulting in indirect CO2 emissions). We follow the license terms for the LLaMa-3.2-1B-Instruct model we use; all models and datasets permit their use as part of the research. As we transform conditional generation into the classification problem (generating only labels), in most cases, we minimize the problem of generating offensive or biased content.

Importantly, in line with the open-science spirit, \fwname is an open-source downstream fork of LLaMA-Factory, licensed under the Apache-2.0 license (we respect the license and add append headers to the files that we have added or modified).

\paragraph{Impact Statement: CO2 Emissions Related to Experiments.} The experiments in this paper require GPU computing resources as we train and evaluate 1 model for different methods (3) and datasets (4). Overall, the experiments, including evaluations (which did not require training but still utilized GPU resources for inference) and preliminary experiments (which are outside the scope of our work), were conducted using a private infrastructure with a carbon efficiency of 0.432 kgCO$_2$eq/kWh. Approximately 50 hours of computation were performed on hardware of type A100 PCIe 40GB (TDP of 250W). Total emissions are estimated to be 9.24 kg CO$_2$eq, of which 0\% were directly offset. Whenever possible, we tried to reduce the computational costs.

\section{Further Details}\label{app:sec:details}

In this section, we include detailed information about \fwname that can be used by advanced users to further understand or extend our framework. In Table \ref{app:tab:peft_comparison}, we provide a comparison of different easy-to-use fine-tuning frameworks in terms of available PEFT methods, highlighting the undeniable contribution of \fwname.

\begin{table*}[t]
\centering
\resizebox{\textwidth}{!}{%
\begin{tabular}{@{}l|cccccccc|ccccc|cccc|cc|c@{}}
\toprule
 & \multicolumn{8}{c|}{Reparametrized} & \multicolumn{5}{c|}{Soft Prompt-Based} & \multicolumn{4}{c|}{Adapter-Based} & \multicolumn{2}{c|}{Selective} & \multirow{2}{*}{PEFT Extensibility} \\ \cmidrule(r){1-20}
 & LoRA & QLoRA & DoRA & LoRA+ & PiSSA & GaLore & OFT & SVFT & Prompt Tuning & Prefix Tuning & P-Tuning & P-Tuning V2 & MTP & IA$^3$ & Bottleneck Adapter & Sequential Bottleneck Adapter & Parallel Adapter & BitFit & LNTuning &  \\ \midrule
\fwname & \ding{52} & \ding{52} & \ding{52} & \ding{52} & \ding{52} & \ding{52} & \ding{52} & \ding{52} & \ding{52} & \ding{52} & \ding{52} & \ding{52} & \ding{52} & \ding{52} & \ding{52} & \ding{52} & \ding{52} & \ding{52} & \ding{52} & \ding{52} \\
LLaMA-Factory & \ding{52} & \ding{52} & \ding{52} & \ding{52} & \ding{52} & \ding{52} & \ding{52} &  &  &  &  &  &  &  &  &  &  &  &  &  \\
FastChat & \ding{52} & \ding{52} &  &  &  & \ding{52} &  &  &  &  &  &  &  &  &  &  &  &  &  &  \\
LitGPT & \ding{52} & \ding{52} &  &  &  &  &  &  &  &  &  &  &  &  &  &  &  &  &  &  \\
LMFlow & \ding{52} & \ding{52} &  &  &  & \ding{52} &  &  &  &  &  &  &  &  &  &  &  &  &  &  \\
Axolotl & \ding{52} & \ding{52} &  &  &  &  &  &  &  &  &  &  &  &  &  &  &  &  &  &  \\
Open-Instruct & \ding{52} & \ding{52} &  &  &  & \ding{52} &  &  &  &  &  &  &  &  &  &  &  &  &  &  \\
H2O LLM Studio & \ding{52} & \ding{52} & \ding{52} &  &  &  &  &  &  &  &  &  &  &  &  &  &  &  &  &  \\ \bottomrule
\end{tabular}%
}
\caption{Comparison of different PEFT methods available in \fwname with popular LLM fine-tuning frameworks. Current frameworks do not include support for other than reparametrized PEFT methods, while most of them are LoRA variations. These are all PEFT methods that were tested for functionality. PEFT Extensibility means that the framework also supports the modular addition of newly created PEFT methods, either by PEFT provider frameworks or directly by users.}
\label{app:tab:peft_comparison}
\end{table*}

\subsection{Preprocessing datasets}\label{app:details:preprocess}
Out of 27 included datasets, we namely preprocess and adapt MultiRC, WiC, COPA, ReCoRD, WSC, MMLU, PIQA, SIQA, HellaSwag, Wingrande, OBQA, MathQA, and SVAMP datasets. We upload all our adapted and preprocessed dataset versions to HuggingFace Hub\footnote{\url{https://hf.co/collections/kinit/peft-factory}}. Additionally, some datasets contain numerical values by default, formatted as \texttt{class values} in the HuggingFace dataset class. We convert such formats to textual representations to ensure compatibility with autoregressive generative models. Therefore, we transform multiple input and output columns of a single dataset to just a two-column format, including only \textit{input} and \textit{output} for the LLM.

\subsection{Efficiency Metrics}\label{app:details:pscp}
The PSCP metric comprises a set of constants that must be configured to function properly. Specifically \textit{pspc\_cp}, \textit{pscp\_cf}, \textit{pscp\_cm}, \textit{pspc\_bp}, \textit{pscp\_bf}, and \textit{pscp\_bm}. The $C$ values are set by the first three attributes, and the $\beta$ values are set by the last three attributes. We also provide default values for these attributes.

The $C$ values in PSCP calculation represent reference constants used for scaling the parameters (pscp\_cp), inference time (pscp\_cf), and peak memory usage (pscp\_cp). The $\beta$ values are defaulty set to 1, but can be set to any positive number. The higher the number, the higher the importance of the number of parameters \textit{pspc\_bp}, inference time \textit{pspc\_bf}, and peak memory usage \textit{pspc\_bm}. For detailed information on how to set these constants and the full equation, see \citet{peftbench}. 

\subsection{Graphical User Interface}
\fwname utilized LLaMA-Board graphical user interface based on Gradio~\citep{abid2019gradio}. In this section, we describe the changes to the graphical user interface that enable fine-tuning LLMs with various PEFT methods.

\begin{figure*}
    \centering
    \includegraphics[width=\textwidth]{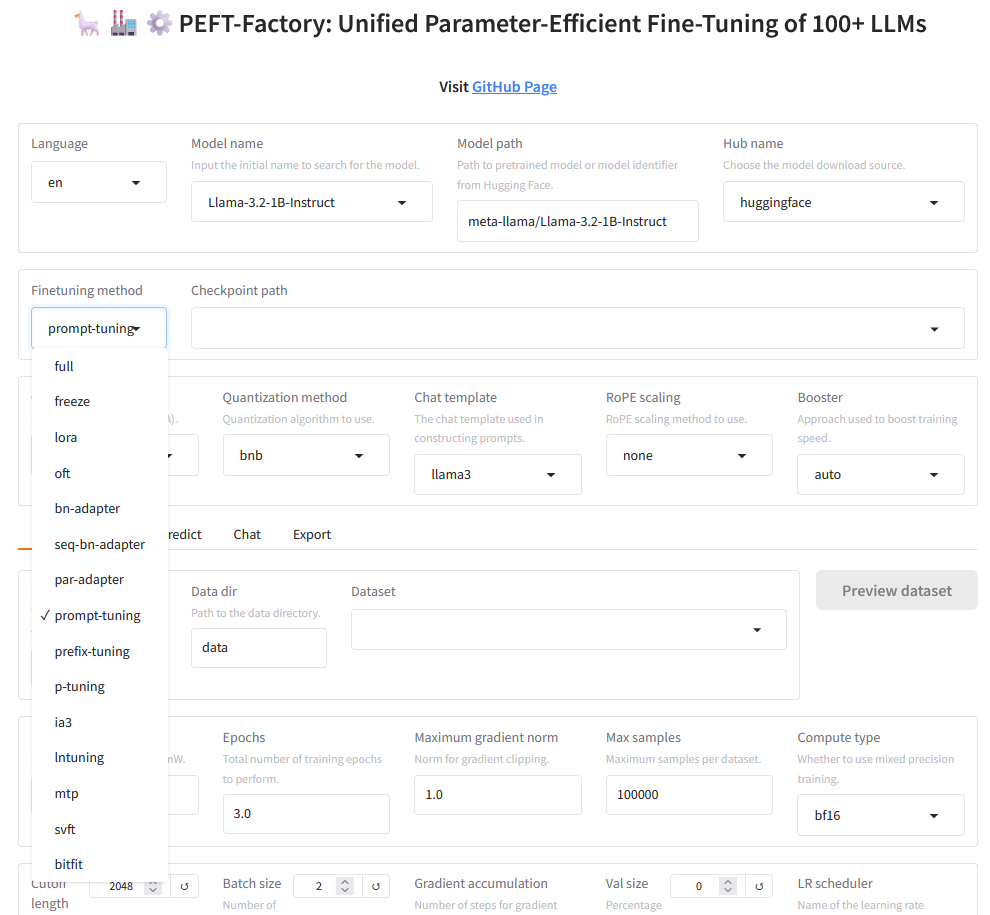}
    \caption{Selection of PEFT methods from \texttt{Finetuning method} dropdown menu. All 19 PEFT methods included in \fwname are available to choose.}
    \label{app:fig:peft_screenshot}
\end{figure*}

\begin{figure*}
    \centering
    \includegraphics[width=\textwidth]{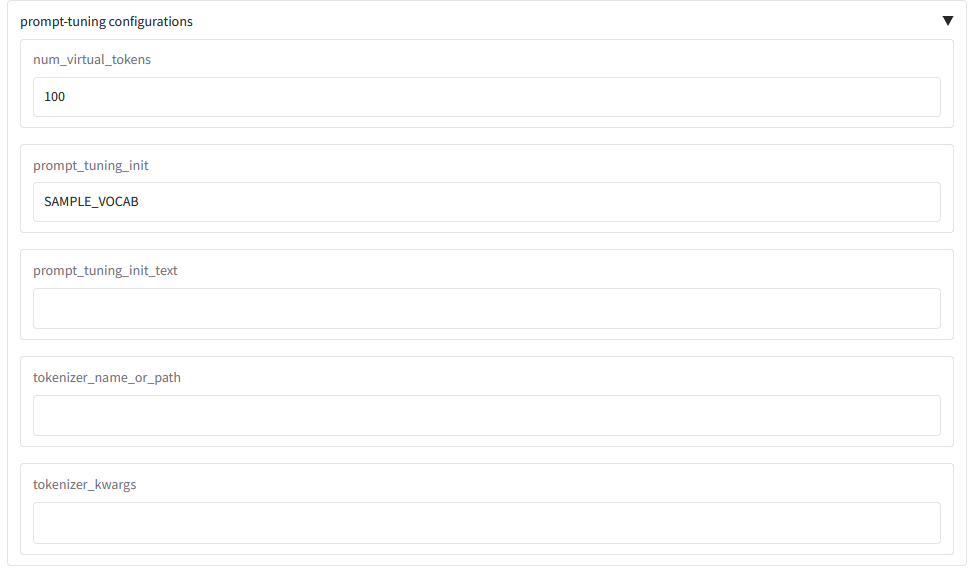}
    \caption{Configuration options for the Prompt Tuning method.}
    \label{app:fig:pt_screenshot}
\end{figure*}

During construction of the Gradio interface, \fwname takes the available PEFT methods and their configurations and constructs an interface for each configuration. Figure \ref{app:fig:peft_screenshot} shows the available PEFT methods to choose from the list. Each PEFT method contains default values that will be set automatically. However, the configuration can be further specified by the detailed configuration shown in Figure \ref{app:fig:pt_screenshot}, which displays the configuration options for the Prompt Tuning method~\cite{lester-etal-2021-power}.

\begin{figure*}
    \centering
    \includegraphics[width=\textwidth]{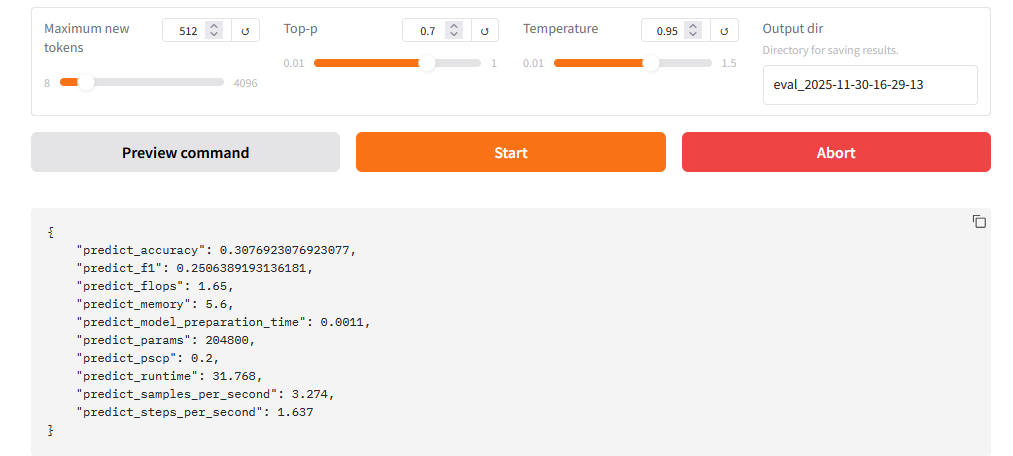}
    \caption{Classification and PSCP results for prediction after training with Prompt Tuning.}
    \label{app:fig:pt_results}
\end{figure*}

\section{Custom PEFT Method Templates}\label{app:sec:template}
We provide minimal templates for the model.py and config.py files to design a PEFT method compatible with the \fwname custom PEFT interface, as documented in our framework\footnote{For detailed information, please visit the \href{https://peft-factory.readthedocs.io/en/latest/contributing.html\#templates}{Templates section} of \fwname documentation.}.

Additionally, we provide an example directory structure (shown in Figure \ref{app:fig:directory}) that can be used to ensure compatibility with dynamic loading of \fwname.

\begin{figure}[H]
  \parbox{\columnwidth}{%
            \centering\small
            \begin{minipage}[t]{\columnwidth}
                \textbf{Custom Method Directory Structure}
                \dirtree{%
                .1 ./peft.
                .2 <CustomMethod>.
                    .3 model.py.
                    .3 config.py.
                }
            \end{minipage}\hfill
}
\caption{Example directory structure of custom PEFT interface used for dynamic loading of PEFT custom methods.}
\label{app:fig:directory}
\end{figure}

\end{document}